\documentclass[pdflatex,sn-mathphys]{sn-jnl}% Math and Physical Sciences Reference Style
\usepackage{makecell}
\usepackage{CJKutf8}
\usepackage[section]{placeins}
\jyear{2021}%

\theoremstyle{thmstyleone}%
\theoremstyle{thmstyletwo}%

\theoremstyle{thmstylethree}%

\begin{document}
\begin{CJK}{UTF8}{gbsn}
\title[An interpretable similar case matching framework]{An interpretable similar case matching framework for legal applications}

%%=============================================================%%
%% Prefix	-> \pfx{Dr}
%% GivenName	-> \fnm{Joergen W.}
%% Particle	-> \spfx{van der} -> surname prefix
%% FamilyName	-> \sur{Ploeg}
%% Suffix	-> \sfx{IV}
%% NatureName	-> \tanm{Poet Laureate} -> Title after name
%% Degrees	-> \dgr{MSc, PhD}
%% \author*[1,2]{\pfx{Dr} \fnm{Joergen W.} \spfx{van der} \sur{Ploeg} \sfx{IV} \tanm{Poet Laureate} 
%%                 \dgr{MSc, PhD}}\email{iauthor@gmail.com}
%%=============================================================%%

\author[1]{\fnm{Nankai} \sur{Lin}}\email{neakail@outlook.com}

\author[2]{\fnm{Haonan} \sur{Liu}}\email{haonan.liu.edu@gmail.com}
% \equalcont{These authors contributed equally to this work.}

\author[1]{\fnm{Jiajun} \sur{Fang}}\email{653639791@qq.com}

\author*[3]{\fnm{Dong} \sur{Zhou}}\email{dongzhou@gdufs.edu.cn}

\author*[1,4]{\fnm{Aimin} \sur{Yang}}\email{amyang@gdut.edu.cn}

% \equalcont{These authors contributed equally to this work.}

\affil[1]{\orgdiv{School of Computer Science and Technology}, \orgname{Guangdong University of Technology}, \orgaddress{ \city{Guangzhou}, \postcode{510006}, \state{Guangdong}, \country{China}}}

\affil[2]{ \orgname{KTH Royal Institute of Technology}, \orgaddress{ \city{Stockholm},  \country{Sweden}}}

\affil[3]{\orgdiv{School of Information Science and Technology}, \orgname{Guangdong University of Foreign Studies}, \orgaddress{ \city{Guangzhou}, \postcode{510006}, \state{Guangdong}, \country{China}}}

\affil[4]{\orgdiv{School of Computer Science and Intelligence Education}, \orgname{Lingnan Normal University}, \orgaddress{ \city{Zhanjiang}, \postcode{524000}, \state{Guangdong}, \country{China}}}

%%==================================%%
%% sample for unstructured abstract %%
%%==================================%%

% \vspace{120 pt}

\abstract{Similar Case Matching (SCM) plays a pivotal role in the legal system by facilitating the efficient identification of similar cases for legal professionals. While previous research has primarily concentrated on enhancing the performance of SCM models, the aspect of interpretability has been overlooked. To bridge this gap, this study proposes an integrated pipeline framework for interpretable SCM. The framework comprises four modules: a judicial feature sentence identification module, a case matching module, a feature sentence alignment module, and a conflict resolution module. In contrast to current SCM methods, our framework extracts feature sentences within a case that contain essential information, conducts case matching based on these extracted features, and aligns the corresponding sentences in two cases to provide evidence of case similarity. In instances where the results of case matching and feature sentence alignment exhibit conflicts, our framework successfully resolves these inconsistencies. The experimental results show the effectiveness of our proposed framework, establishing a new benchmark for interpretable SCM.}


%%\pacs[JEL Classification]{D8, H51}

%%\pacs[MSC Classification]{35A01, 65L10, 65L12, 65L20, 65L70}

\maketitle

\section{Introduction}
With the development of information technology, more and more traditional industries are benefiting from artificial intelligence. Legal AI has become a trendy area of research and, as such, has received much more attention from legal professionals and AI researchers \citep{10.1145/3512898,schwemer2021legal,wang2022intelligent}.

The task of the SCM aims to detect whether the two given cases are similar or not. In a standard legal system, the judgement of a case is influenced by the most similar cases in the past. However, in the traditional administration of justice, legal professionals spend a great deal of time and effort searching for similar cases to provide them with the necessary knowledge and evidential support in court. Therefore, the automatic retrieval of similar cases is of great practical value and relevance as it reduces the heavy workload of legal professionals.

Since the release of the Chinese SCM task in CAIL (Chinese AI and Law Challenge) 2019\footnote{http://cail.cipsc.org.cn/} \citep{xiao2019cail2019}, scholars have started to focus on the performance of SCM on Chinese, and many valuable studies have emerged. However, these efforts focus only on whether the two cases are similar but do not explain for the result.

The lack of explainability in previous studies can lead to a risky society of algorithmic discrimination, algorithmic killing, and "information cocoon", which need to be solved urgently. Due to the fact that the current legal system preserves the neutral value of technological instruments while ignoring technological legal requirements, it is impossible for unjustified algorithms to control and prevent social problems. As the artificial intelligence become "smarter", the explainable AI is an emerging field targeting at solving the potential social risks. The field promotes use to reconsider the usage of AI application in many fields including in the legal system. So the interpretability of Legal AI needs to be fully valued and considered \citep{lin2020legal,yanhong2020research,9521221}. 

This paper addresses the issue of interpretability in SCM tasks and presents an interpretable pipeline framework. The framework consists of four modules: a judicial feature sentence identification module, a case matching module, a feature sentence alignment module, and a conflict resolution module. In contrast to existing SCM methods, our framework focuses on identifying feature sentences in a case that contain crucial information, conducting similar case matching based on the extracted feature sentence results, and aligning the feature sentences in the two cases to provide evidence for case similarity. As similar case matching results might conflict with feature sentence alignment results, our framework includes a mechanism for resolving inconsistent outcomes. Experimental results demonstrate the effectiveness of our proposed framework, and its modular design allows for easy integration of new components and techniques to enhance model performance.

In summary, the contributions of this paper are:

(1) The paper proposes a framework oriented towards SCM, which can effectively retrieve similar cases with explanation.

(2) The framework proposed in this paper contains several modules, each of which is complementary in function. The framework is flexible. Adjustments to each module will not affect other modules. Each module can be replaced independently, making it easy to improve.

(3) On the Chinese interpretable similarity case matching task, the framework proposed in this paper outperforms the baseline method by 33.04\%, providing a new benchmark for this task.

\begin{figure}
  \centering
  \includegraphics[width=1\textwidth]{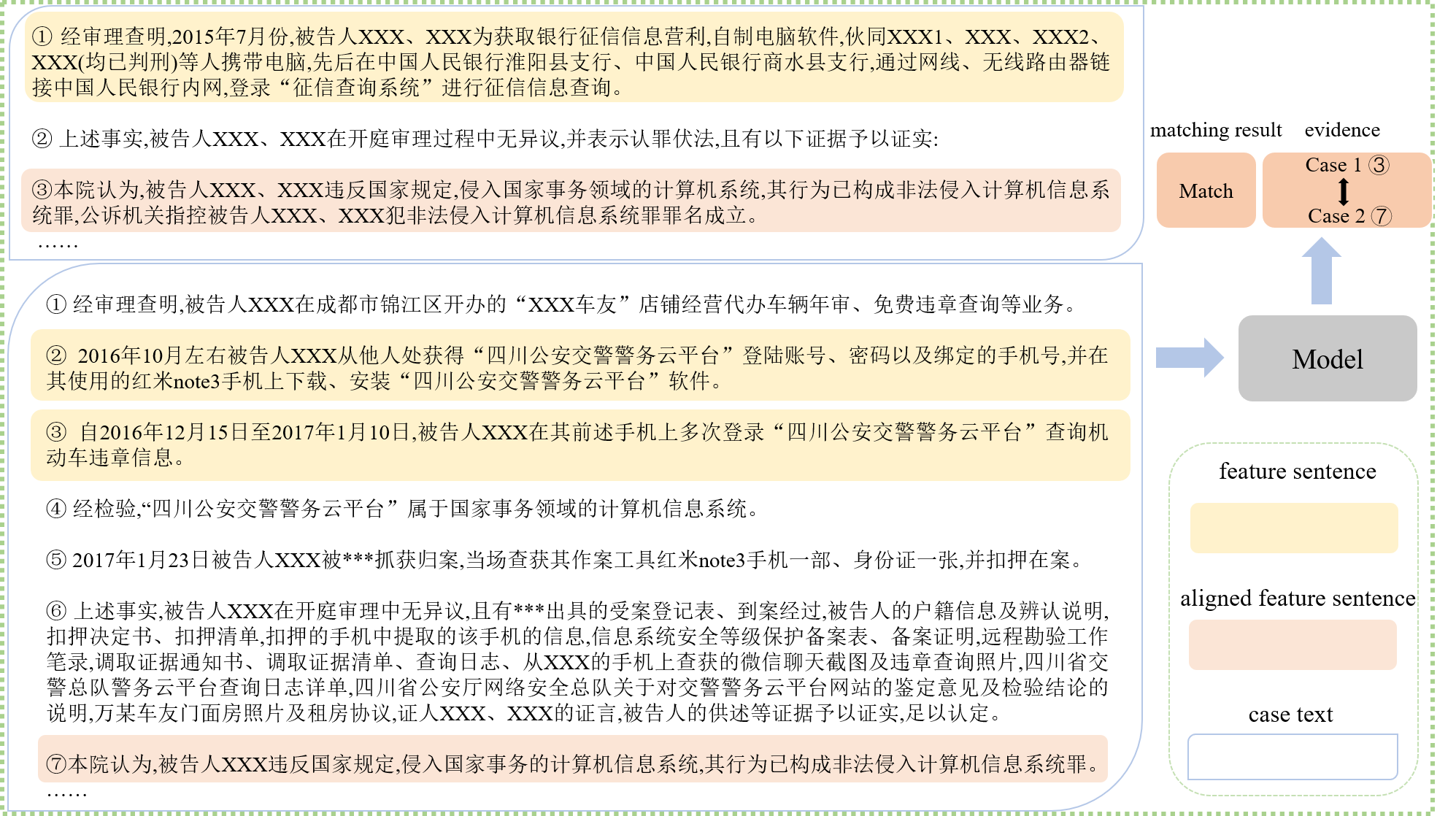}
  \caption{Task Definition.} 
  \label{fig:0} 
\end{figure}

\section{Related Work}
In judicial decisions, legal professionals often make decisions based on past cases. Therefore, identifying similar cases is the primary concern in the judgment. Similar Case Matching (SCM) has emerged as a crucial area of study for LegalAI \citep{zhong-etal-2020-nlp}. It focuses on identifying pairs of similar cases, and there are many ways to define the similarity. SCM calls for modeling the link between cases using data at various granularity levels, such as the fact, event, and element levels. In other words, SCM is a specific type of semantic matching that can help in retrieving legal knowledge \citep{xiao2018cail2018}. 

With the emergence of LegalIR datasets such as COLIEE \citep{kano2018coliee}, CaseLaw \citep{10.1145/3209978.3210161}, and CM \citep{xiao2019cail2019}, SCM has gradually attracted the attention of scholars. Benchmarks for the studies of LegalIR are provided by these datasets. The issues of semantic text matching are addressed using a variety of deep learning models. In order to attain the best results on COLIEE, Tran, Nguyen and Satoh \cite{10.1145/3322640.3326740} suggested a CNN-based model with document and sentence level pooling. Other researchers are investigating the use of better embedding techniques for LegalIR \citep{10.1007/978-3-030-01177-2_12}. In order to improve the model’s performance in the primary task of similar case matching, Peng, Yang and Lu \cite{PENG2020106514} proposed a multi-task learning framework with “de- and re-construction”, which makes use of the extraction of sub-tasks based on sentence-level knowledge components to improve the document level representation. The semantic text matching model’s legal feature vector is added by Hong, Zhou, Zhang, Li and Mo \cite{9207528}, and BERT is used as the encoding layer to capture distant dependencies in the case documents. Li, Lu, Le and He \cite{li2022iacn} developed a brand-new interactive attention capsule network model. It attempts to mimic the method of expert legal judgment, which captures the similarity of minute details to produce an understandable verdict. Additionally, they developed an interactive dynamic routing technique that outperforms the standard dynamic routing in learning the interaction representation of legal aspects among cases.

Although scholars have researched similar case matching, there is still a lack of explanation for the case matching results. To better support the case matching result, we propose a pipeline framework for interpretable SCM.

\section{Task Definition}

Figure 1 shows the task flow of the interpretable similar case matching task. This task is given two cases and asks the model to judge how similar the two cases are (not match, partial match, and match). At the same time, the model needs to identify alignable feature sentences in the cases as evidence support for the two cases that the model considers similar. It is worth noting that the feature sentences contained in a case are not necessarily all alignable. There are three feature sentences: (1) The critical issues about the evidential facts and the application of law disputes, that is, the focus of the dispute, summarized by the judge. (2) Essential facts, such sentences contain abstract words or phrases of legal concepts. (3) Case descriptions, that is, a description of the circumstances of the case. Since the length of the case text is too long, directly matching the text and extracting the evidence will generate many calculations and reduce the model's efficiency. Therefore, this paper attempts to decompose the matching task of similar cases with interpretability. First, we identify all the feature sentences of each case. Then, based on the identified feature sentences, the similarity degree is judged and aligned with the feature sentences.

\section{A pipeline framework for interpretable similarity case matching}
This paper introduces an interpretability framework for similar case matching. In contrast to existing methods, our framework is designed to identify the feature sentence that contains crucial information within a case (detailed in subsection 4.1). Leveraging the extracted feature sentence results, we then proceed to match similar cases (detailed in subsection 4.2) and align the feature sentences from the two cases (detailed in subsection 4.3) to provide evidence supporting the similarity of the cases. To address potential conflicts between the results of the case matching module and the feature sentence alignment module, our framework includes a mechanism to resolve these inconsistencies (detailed in subsection 4.4). The overall proposed framework is visually depicted in Figure 2.
\begin{figure}
  \centering
  \includegraphics[width=1\textwidth]{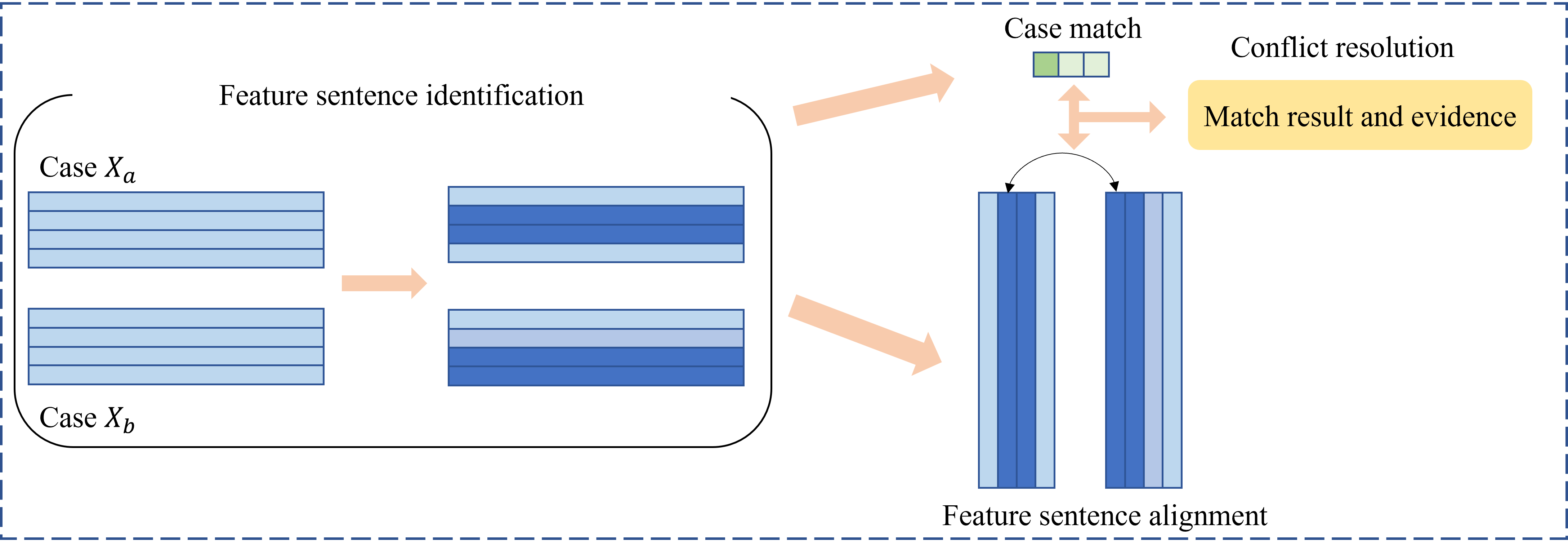}
  \caption{The proposed interpretable similarity case matching framework.} 
  \label{fig:1} 
\end{figure}

\subsection{Feature sentence identification module}
Given a case as the input of the module: $X=[(s_1,y_1 ),(s_2,y_2 ),…(s_n,y_n )]$, $n$ is the number of sentences in the case $X$, $s_i (i \in [1,n])$ represents the $i$-th sentence in the case and $y_i$ is the associated label of it. The label is a binary variable $y \in \{0,1\}$ that indicates whether a sentence is a feature sentence of the case or not. Given a sentence $s_i$, the feature sentence identification module first represents the sentence into a semantic vector using the RoBERTa \citep{liu2019roberta} pre-trained model:
\begin{equation}
    h_i=RoBERTa(s_i)
\end{equation}

Then the vector is fed to a feed-forward network followed by a softmax function to output a label probability distribution:
\begin{equation}
    p_i=softmax(W_{fsi} \cdot h_i+b_{fsi} )
\end{equation}

where $W_{fsi}$ and $b_{fsi}$ are learnable parameters. Notice that the RoBERTa model can be replaced by any pre-trained language models for sentence representation, and the feed-forward network can also be substituted with other complex classifiers. Since the project aims to prove our framework's effectiveness, we keep the components simple and offer flexibility for other possibilities.

\subsection{Case matching module}

The case matching module can classify whether or not two input cases are matched. Given two cases $X_a$ and $X_b$, each contains all the sentences of a case and a label $Y \in \{0,1,2\}$ indicating not match, partial match, and match, respectively. The two cases are first fed to the feature sentence identification module described in the last subsection. The module will identify feature sentences in a case and form a new set of sentences representing the case. Thus, cases $X_a$ and $X_b$ will be transformed into a set of m feature sentences representing $X_a^{'} =[s_{a1},s_{a2},…s_{am}]$ and a set of $l$ feature sentences representing $X_b^{'} =[s_{b1},s_{b2},…s_{bl}]$. With $X_a^{'}$ and $X_b^{'}$ as input, the module first encodes sentences in each case using RoBERTa model and then performs matching in two different modes, as follows.

\subsubsection{Concat Mode}
Under this mode, sentence pairs $(s_{ai},s_{bj})$ from the two cases are concatenated and are fed into the RoBERTa model:
\begin{equation}
    input_{a_i,b_j} =[CLS] s_{ai} [SEP] s_{bj} [SEP]
\end{equation}

The semantic vector then can be retrieved from the RoBERTa model:
\begin{equation}
    h_{a_i,b_j}=RoBERTa(input_{a_i,b_j})
\end{equation}

Then the vector is fed to a feed\-forward network followed by a softmax function to output a label probability distribution：
\begin{equation}
    p_{a_i,b_j}=softmax(W_{cm} \cdot h_{a_i,b_j}+b_{cm} )
\end{equation}

where $W_{cm}$ and $b_{cm}$ are learnable parameters.
\subsubsection{Siamese Mode}
Unlike in the Concat mode, we represent each sentence in the sentence pairs $(s_{ai},s_{bj})$ using RoBERTa model:
\begin{equation}
    h_{a_i} =RoBERTa(s_{a_i})
\end{equation}
\begin{equation}
    h_{b_j} =RoBERTa(s_{b_j})
\end{equation}

We construct a hidden vector by concatenating the two text representations and an absolute difference of the representations. Then it is fed to a feed-forward network followed by a softmax function to output a label probability distribution:
\begin{equation}
    h_{fe}=[h_{a_i};h_{b_j}; \vert h_{a_i} - 
 h_{b_j} \vert ]
\end{equation}
\begin{equation}
    p_{fe}=softmax(W_{fe} \cdot h_{fe}+b_{fe} )
\end{equation}

where $W_{fe}$ and $b_{fe}$ are learnable parameters.

\subsection{Feature sentence alignment module}
For the two cases $X_a^{'}=[s_{a1},s_{a2},…s_{am}]$ and $X_b^{'}=[s_{b1},s_{b2},…s_{bl}]$ after feature identification and filtering, we can form sentence-pairs set $P=[(s_{a1},s_{b1},c_{11}),(s_{a1},s_{b2},c_{12}),…,(s_{am},s_{bl},c_{ml})]$ for alignment, and $P \in R^{m*l}$. For each sentence pair $(s_{ai},s_{bj})$, $i \in [1,m]$, $j \in [1,l]$ there is an associated label $c_{ij} \in \{0,1\}$ indicating whether the two sentences can be aligned or not . The sentence alignment module uses the same two matching modes as in the case matching modules to calculate the alignment probability of a sentence pair. In the testing stage, we first calculate alignment probabilities for each sentence pair in $P$ and get a set of alignment probabilities $p=[p_{00},p_{01},…p_{ml}]$. Then we only preserve sentence pairs with corresponding alignment probabilities larger than 0.5 to form an aligned feature sentence pair set $P_{align}$. The set can explain the case-matching prediction. 
\subsection{Conflict Resolution Module}
Ambiguation between the case matching module and the feature sentence alignment module may happen. Two cases might be predicted as "match" or "partially match" in the case matching module but cannot find aligned sentence pairs in the feature alignment module. To resolve the ambiguity, in this case, we modify the matching label of the two cases to "not match".

\begin{figure}
  \centering
  \includegraphics[width=0.8\textwidth]{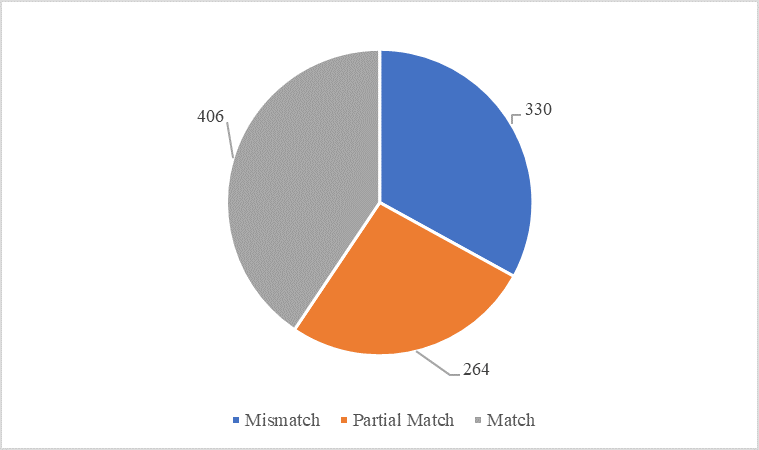}
  \caption{The label distribution of case matching.} 
  \label{fig:2} 
\end{figure}

\section{Experimental data}
The experimental data used in this paper comes from the CAIL 2022 Interpretable Similar Case Matching dataset\footnote{http://cail.cipsc.org.cn/task7.html?raceID=7\&cail\_tag=2022}, which provides a total of 1000 labeled samples for the competition. For each given pair of cases, the matching degree, all feature sentences of the cases, and the aligned feature sentences have been annotated. The matching labels statistics are shown in Figure 3. It can be observed that the weight of the three categories in the dataset is relatively close, and there is no category imbalance. In addition, we further analyzed the distribution of the number of feature sentences for the cases, and the results are presented in Figure 4. The first line in the x-axis indicates the number of feature sentences contained in one case, and the second line indicates the total number of cases with the specific number of feature sentences (2 to 14+). It can be seen that the number of feature sentences in most cases is 3 to 7 sentences. We performed five-fold cross-validation on the dataset. Each fold's samples were divided in equal proportions according to the proportion of labels to construct the training and test sets.

\begin{figure}
  \centering
  \includegraphics[width=1\textwidth]{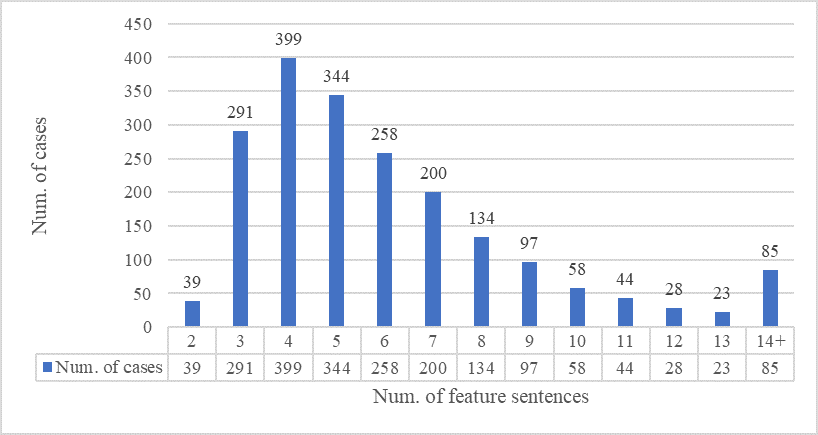}
  \caption{The distribution of the number of feature sentences.} 
  \label{fig:3} 
\end{figure}

Based on the above five-fold dataset, we further construct the data for feature sentence identification and feature sentence alignment. Taking the first fold data as an example, we split all the cases in the first fold (each sample has two cases) into new samples in sentences to construct the training set for feature sentence identification of the first fold. Also, we treat a sample with two cases of standard feature sentence matches that exist in the actual set of sentence pairs supporting the match as a positive sample and vice versa as a negative sample. The distribution of categories in the five-fold training set for feature sentence identification and feature sentence alignment is shown in figure 5 with figure 6. In Figure 5, the tag "Num. of feature sentences" represents the number of feature sentences, and the tag "Num. of non feature sentences" represents the data of non feature sentences. In Figure 6, the tag "aligned" represents that the sentence pair is an aligned feature sentence pair, while the tag "unaligned" represents that the sentence pair is an unaligned sentence pair.

\begin{figure}
  \centering
  \includegraphics[width=1\textwidth]{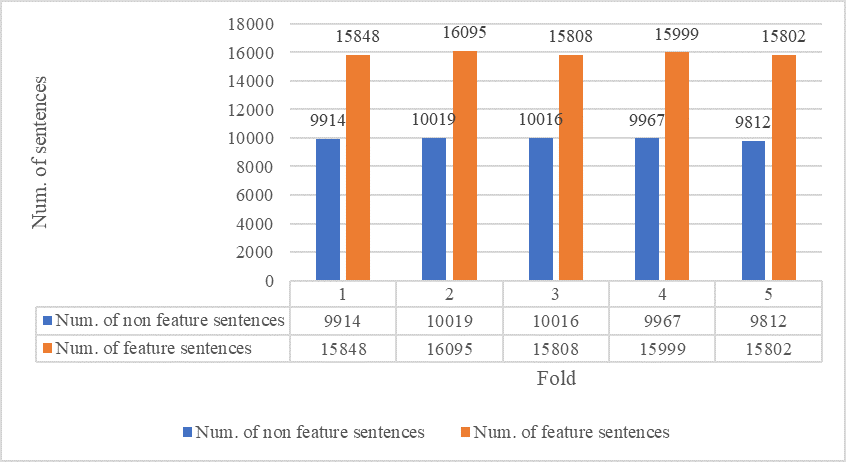}
  \caption{The label distribution of feature sentence identification.} 
  \label{fig:4} 
\end{figure}

\begin{figure}
  \centering
  \includegraphics[width=1\textwidth]{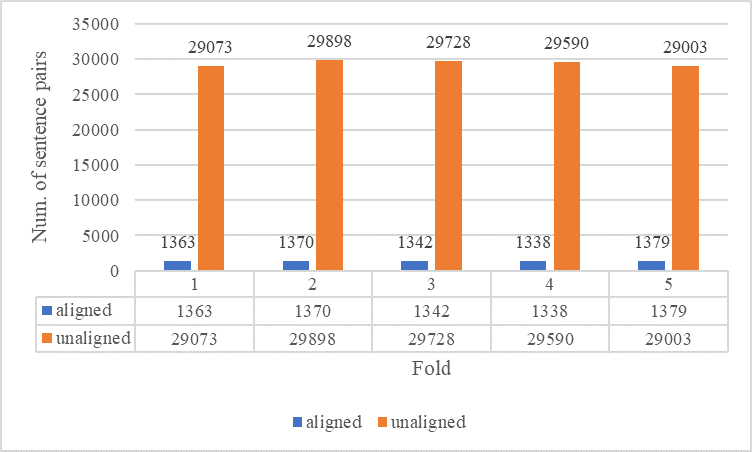}
  \caption{The label distribution of feature sentence alignment.} 
  \label{fig:5} 
\end{figure}

\section{Evaluation Metrics}

For the entire interpretable SCM task, the mean of the matching result score and the interpretation result score is used as the final score. We used Macro-F1 in the classification task as the evaluation metric when calculating the matching result score:
\begin{equation}
F_{sm}=  \frac{(F_{match}+F_{partialmatch}+{F_mismatch})}{3}
\end{equation}
\begin{equation}
\begin{split}
    F_{category} =  \frac{(2 * P_{category} * R_{category})}
    {(P_{category} + R_{category} )} \\
   (category \in \{ match,partialmatch,mismatch\})
\end{split}
\end{equation}
For the interpretation results, we first evaluate the accuracy of feature sentence extraction for individual cases, i.e., we assess how well the judicial feature sentences extracted by the model match the annotated feature sentences. We use Macro-F1 as the evaluation metric, where $S_i^p$ is the set of judicial feature sentences of a single case output by the model, and $S_i^g$ is the set of true feature sentences:
\begin{equation}
    F_{fsi}=  \frac{1}{N} \sum_{i=1}^N (2*\frac{P_i * R_i}{P_i + R_i})
\end{equation}

\begin{equation}
    P_i=  \frac{\vert S_i^p \cap S_i^g \vert }{\vert S_i^p \vert}
\end{equation}

\begin{equation}
    R_i=  \frac{\vert S_i^p \cap S_i^g \vert }{\vert S_i^g \vert}
\end{equation}

Meanwhile, for the case pairs with matching or partial matching labels, the accuracy of feature sentence identification and alignment in the case pairs supporting matching results is evaluated, again using Macro-F1 as the evaluation metric. $S_i^p$ is the set of sentence pairs supporting matching output by the model, and $S_i^g$ is the set of true sentence pairs supporting matching.
\begin{equation}
    F_{fsa}=  \frac{1}{N} \sum_{i=1}^N (2*\frac{P_i * R_i}{P_i + R_i})
\end{equation}
\begin{equation}
    P_i=  \frac{\vert S_i^p \cap S_i^g \vert }{\vert S_i^p \vert}
\end{equation}
\begin{equation}
    R_i=  \frac{\vert S_i^p \cap S_i^g \vert }{\vert S_i^g \vert}
\end{equation}

For the entire interpretable SCM task, the final score is:
\begin{equation}
    F_{final}=  \frac{1}{2} \cdot F_{sm}+  \frac{1}{4} \cdot (F_{fsi}  + F_{fsa})
\end{equation}

In evaluating the performance of the judicial feature sentence recognition module, the case matching module, and the feature sentence alignment module, respectively, Macro-F1 was used as the evaluation metric.

\section{Experiments}

\subsection{Experimental setup}
All experiments were carried out using PyTorch\footnote{https://github.com/pytorch/pytorch} and an RTX 8000 with 48 GB of memory. We build our framework based on Transformers\footnote{https://github.com/huggingface/transformers}. Furthermore, we choose the RoBERTa-large\footnote{https://huggingface.co/hfl/chinese-roberta-wwm-ext-large} \citep{liu2019roberta} model as the pre-trained model adopted by our framework. As can be seen from section 4, the data distribution of the feature sentence alignment module is extremely unbalanced, so we use the label weight strategy \citep{Yan_2017_CVPR} in this module, where the loss weights of negative samples and positive samples are set to 0.1 and 1, respectively. In order to verify the ease of modification of our framework, we tried to add FGM strategies to each module. A brand-new regularization technique called FGM \citep{goodfellow2014explaining,sato2018interpretable,miyato2016adversarial} enhances the robustness of misclassifying minor perturbed inputs. We also compare our framework with the baseline model\footnote{https://github.com/china-ai-law-challenge/CAIL2022/tree/main/kjslapp} provided by the competition
for both accuracy and interpretability. We aim to propose a stronger benchmark than the existing baseline method. In this way, our framework provides a more appropriate comparison method for the follow-up research on the interpretable cases matching. Table 2 summarises the hyper-parameters used in our experiments.
\begin{table}[h]
\centering
\caption{Hyper-parameter values.}
\begin{tabular}{cc}
\hline
Parameter & Value \\
\hline
Feature dimension & 1024 \\
Batch size & 4 \\
Dropout & 0.5 \\
Number of epochs & 10 \\
Learning rate & 5e-6 \\
Optimizer & Adam \\
Max length in feature sentence identification module & 128 \\
Max length in case matching module & 512 \\
Max length in feature sentence alignment module & 128 \\
\hline
\end{tabular}
\label{tab:accents}
\end{table}

\begin{table}[h]
\centering
\caption{Experimental results of our framework.}
\begin{tabular}{cccc}
\hline
Method & \makecell{Matching \\ result score} & \makecell{Interpretation \\ result score} & Final score \\
\hline
Baseline & 65.38 & 22.37 & 43.87 \\
Our Framework (Concat) & 67.85 & 81.55 & 74.70 \\
Our Framework (Siamese) & 68.72 & 79.04 & 73.88 \\
Our Framework (Concat and FGM) & \textbf{71.74} & \textbf{82.07} & \textbf{76.91} \\
Our Framework (Siamese and FGM)) & 69.21 & 79.09 & 74.15 \\
\hline
\end{tabular}
\label{tab:accents}
\end{table}

\subsection{Experimental results}

\subsubsection{Main results}
We employ four combinations for the interpretable similar case matching task under our proposed framework. The results are shown in Table 2. It can be seen that, based on different combinations, our framework can exceed the performance of the existing baseline model by 30\%. Among them, when the case matching module and the case matching module use the "concat" mode, and all modules use the FGM strategy, our framework achieves the optimal performance, and the final score reaches 76.91\%. We can easily apply different models in different modules, showing that our framework is straightforward to operate and modify.

\begin{table}[h]
\centering
\caption{Experimental results of feature sentence identification module.}
\begin{tabular}{cc}
\hline
Method & Macro-F1 \\
\hline
RoBERTa & 91.80 \\
RoBERTa with FGM & \textbf{92.08} \\
\hline
\end{tabular}
\label{tab:accents}
\end{table}

\subsubsection{Each modules' results}
We further explore the application of different models under different modules, and the experimental results are shown in Table 3, Table 4, and Table 5. As can be seen from Table 2, the feature sentence identification is an easy-to-learn task. Even without adding other strategies, feature sentence identification module can achieve good performance, and the Macro-F1 value reaches 91.80\%. Existing strategies, such as FGM, have improved the module, and the Macro-F1 value has only increased by 0.28\%. Therefore, we suggest that follow-up research should focus on improving the performance of other modules.

\begin{table}[h]
\centering
\caption{Experimental results of case matching module.}
\begin{tabular}{ccc}
\hline
Method & Used data & Macro-F1 \\
\hline
Matching & Full text & 63.17 \\
Matching with FGM & Full text & 63.61 \\
Matching & Feature sentences & 63.76 \\
Concat & Feature sentences & 67.97 \\
Concat with FGM & Feature sentences & \textbf{70.26} \\
Siamese & Feature sentences & 68.15 \\
Siamese with FGM & Feature sentences & 69.17 \\
\hline
\end{tabular}
\label{tab:accents}
\end{table}

\begin{table}[h]
\centering
\caption{Experimental results of feature sentence alignment module.}
\begin{tabular}{cc}
\hline
Method & Macro-F1 \\
\hline
Matching & 68.04 \\
Concat & 86.07 \\
Concat and FGM & \textbf{86.55} \\
Siamese & 83.39 \\
Siamese and FGM & 83.50 \\
\hline
\end{tabular}
\label{tab:accents}
\end{table}

Table 4 shows the performance of different strategies on the case matching module. In addition to the two matching modes mentioned in subsection 3.2, we also use a similar version of the "siamese" mode, the "matching" mode; that is, the text representations of the two texts are directly added without additional feature extraction. In the case matching module, we can see that using the feature sentence identification module to extract the feature sentences for constructing the input text performs better than directly using all the case information as the input text. This shows the effectiveness and structural soundness of our pipeline framework. Among them, the case matching module combining FGM strategy and "concat" mode achieved the best performance, and the Macro-F1 value reached 70.26\%. Similarly, the combination also achieved the best performance in the feature sentence alignment module, and the experimental results are shown in Table 5.

\begin{table}[h]
\centering
\caption{Ablation study.}
\begin{tabular}{cccc}
\hline
Method & \makecell{Matching \\ result score} & \makecell{Interpretation \\ result score} & Final score \\
\hline
Our Framework (Concat) & 67.85 & 81.55 & 74.70 \\
- Conflict Resolution Module & 67.10 & 81.55 & 74.32 \\
\hline
Our Framework (Siamese) & 68.72 & 79.04 & 73.88 \\
- Conflict Resolution Module & 67.18 & 79.04 & 73.11 \\
\hline
Our Framework (Concat and FGM) & 71.74 & 82.07 & 76.91 \\
- Conflict Resolution Module & 70.60 & 82.07 & 76.33 \\
\hline
Our Framework (Siamese and FGM)) & 69.21 & 79.09 & 74.15 \\
- Conflict Resolution Module & 68.53 & 79.09 & 73.81 \\
\hline
\end{tabular}
\label{tab:accents}
\end{table}

\subsection{Ablation study}
Since the conflict resolution module is not necessary for our framework, we conducted an ablation study on it, and the experimental results are shown in Table 6. It can be seen that under different models, the conflict resolution module can bring specific improvements. Through a simple resolution strategy, the framework's performance can be improved by 0.34\% to 0.77\%, indicating that the design of the conflict resolution module is meaningful for our pipeline framework.

%% Loading bibliography style file
% \bibliographystyle{model1-num-names}

\begin{table}[h]
\footnotesize
\centering
\caption{Case study.}
\begin{tabular}{p{3.8cm}p{7.4cm}}
\hline
Case 1 & ①	经审理查明:2019年7月12日,被告人XXX通过微信朋友圈发布招募卖血人员信息,后于7月15日7时在北京市通州区北运河西地铁站组织5人前往通州区某街道政务服务中心二层进行有偿献血;同年7月15日12时许,被告人XXX在北京市昌平区某大学附近被民警抓获;被告人XXX作案时所用手机已扣押。

②	上述事实,有证人XXX1、高某、任某、XXX、XXX、XXX、XXX2的证言,被告人XXX的供述,扣押笔录、辨认笔录,视听资料,微信聊天记录截图,司法鉴定意见书,接报案经过、到案经过、破案报告、北京市无偿献血登记表、扣押决定书、扣押清单、户籍信息查询单、电话查询记录等证据证实,足以认定。

③	本院认为,被告人XXX无视法律,非法组织他人卖血,其行为已构成非法组织卖血罪,依法应予惩处。

④	公诉机关指控的罪名成立。

⑤	被告人XXX到案后如实供述自己的犯罪事实,自愿认罪认罚,依法可以从轻处罚。

⑥	辩护人XXX的辩护意见经查属实且与法有据,本院予以采纳 \\
\hline
Case 2 & ① 经审理查明:被告人文建于2019年10月18日,在北京市朝阳区国贸等地非法组织XXX、XXX、XXX、XXX、XXX某等人进行卖血,于10月21日在北京市顺义区第二医院非法组织XXX进行卖血,获利共计人民币700元。

② 后XXX被***查获归案。

③ ***起获了XXX的vivo牌手机1部,另起获了手机3部、笔记本1本、银行卡8张,现扣押在案。

④ 上述事实,被告人文建及辩护人在庭审过程中未提出异议,且有证人XXX等人的证言、辨认笔录、献血人名单等证以证实,足以认定。

本院认为,被告人文建无视国法,非法组织他人出卖血液,其行为触犯了刑法,已构成非法组织卖血罪,依法应予惩处。

⑤ 北京市朝阳区人民检察院指控被告人文建犯非法组织卖血罪的事实清楚,证据确实、充分,指控罪名成立。

⑥ 被告人文建归案后如实供述自己的罪行,当庭认罪认罚,故本院对其所犯罪行依法予以从轻处罚。

⑦ 辩护人关于文建获利不多、认罪认罚及建议法庭对其从轻处罚的辩护意见,本院予以采纳;其他辩护意见,缺少事实或者法律依据,本院不予采纳。

⑧ 辩护人文建的犯罪所得,依法应予没收。

⑨ 在案之vivo牌手机,系文建的犯罪工具,依法应予没收;在案之其他物品,和指控事实无关,依法退回公诉机关。

⑩ 综上,根据被告人文建犯罪的事实、犯罪的性质、情节以及对于社会的危害程度,本院依照《中华人民共和国刑法》第三百三十三条第一款、第六十一条、第六十七条第三款、第四十五条、第四十七条、第五十二条、第五十三条、第六十四条及《中华人民共和国刑事诉讼法》第十五条之规定,判决如下: \\
\hline
Gold match label & matching \\
\hline
Gold feature sentences of case 1 & ①, ③, ⑤ \\
\hline
Gold feature sentences of case 2 & ①, ⑤, ⑦, ⑧ \\
\hline
Gold aligned feature sentences &　①-①， ③-⑤， ⑤-⑦ \\
\hline
Predicted match label by our framework & matching \\
\hline
Predicted feature sentences of case 1 by our framework & ①, ③, ⑤ \\
\hline
Predicted feature sentences of case 2 by our framework & ①, ⑤, ⑦, ⑧ \\
\hline
Predicted aligned feature sentences by our framework &　①-①， ③-⑤， ⑤-⑦ \\
\hline
Predicted match label by baseline method & matching \\
\hline
Predicted feature sentences of case 1 by baseline method & ①, ②, ③ \\
\hline
Predicted feature sentences of case 2 by baseline method & ①, ②, ③ \\
\hline
Predicted aligned feature sentences by baseline method &　①-①， ②-②， ③-③ \\
\hline

\end{tabular}
\label{tab:accents}
\end{table}

\subsection{Case study}

We further conduct a case study to demonstrate the superior performance of our method over existing baseline models. The results of the case study are shown in Table 7. For the given example, both our method and the baseline model obtain correct predictions for the matching results of the identified cases. In terms of providing interpretability evidence, we show our superior performance. We not only correctly identify all the feature sentences in the two cases but also correctly align the feature sentences. However, the baseline method only correctly recognizes some feature sentences, and the wrong recognition results further affect the performance of subsequent alignment methods. It can be seen that our method can not only accurately complete the case-matching task but also provide high-quality supporting evidence.

\section{Conclusion}
Existing SCM research has focused on improving the model’s performance but not on its interpretability. Therefore, this paper proposes a pipeline framework for interpretable SCM. The experimental results show the effectiveness of our framework, and our work provides a new benchmark for interpretable SCM. In addition, we suggest that follow-up research should focus on improving the performance of the case matching and feature sentence alignment modules.

\section{Acknowledgement}
This work was supported by the Guangdong Basic and Applied Basic Research Foundation of China (No. 2023A1515012718) and the Philosophy and Social Sciences 14th Five-Year Plan Project of Guangdong Province (No. GD23CTS03).

\section{Declarations}
\subsection{Conflicts of interests}
The authors declare that we do not have any commercial or associative interest that represents a conflict of interest in connection with the work submitted and that the research do not involve human participants and/or animals.
\subsection{Data availability}
Data will be made available on request.

\bibliography{sn-bibliography}% common bib file
%% if required, the content of .bbl file can be included here once bbl is generated
%%\input sn-article.bbl

%% Default %%
%%\input sn-sample-bib.tex%

\end{CJK}
\end{document}